\newif\ifemnlpsubmission
\ifdefined\emnlpsubmission
\emnlpsubmissiontrue
\else
\emnlpsubmissionfalse
\fi

\documentclass[11pt]{article}

\ifemnlpsubmission
\usepackage[review]{acl}
\else
\usepackage[preprint]{acl}
\fi

\usepackage{times}
\usepackage{latexsym}
\usepackage[T1]{fontenc}
\usepackage[utf8]{inputenc}
\usepackage{microtype}
\usepackage{inconsolata}

\usepackage{graphicx}
\usepackage{booktabs}
\usepackage{array}
\usepackage{multirow}
\usepackage{amsmath}
\usepackage{amssymb}
\usepackage{bm}
\usepackage{xcolor}
\usepackage{enumitem}
\usepackage{xspace}
\usepackage{url}
\usepackage{algorithm}
\usepackage{algpseudocode}
\usepackage{float}
\usepackage{placeins}

\hypersetup{
  pdftitle={SkillHone: A Harness for Continual Agent Skill Evolution Through Persistent Decision History},
  pdfauthor={Zhiwei Li and Yong Hu}
}

\definecolor{wechatgreen}{HTML}{3D9B7A}
\ifemnlpsubmission\else
\AddToHookNext{shipout/foreground}{%
  \put(0,-48){\makebox[\paperwidth][l]{\hspace*{2.5cm}%
    \includegraphics[width=90pt]{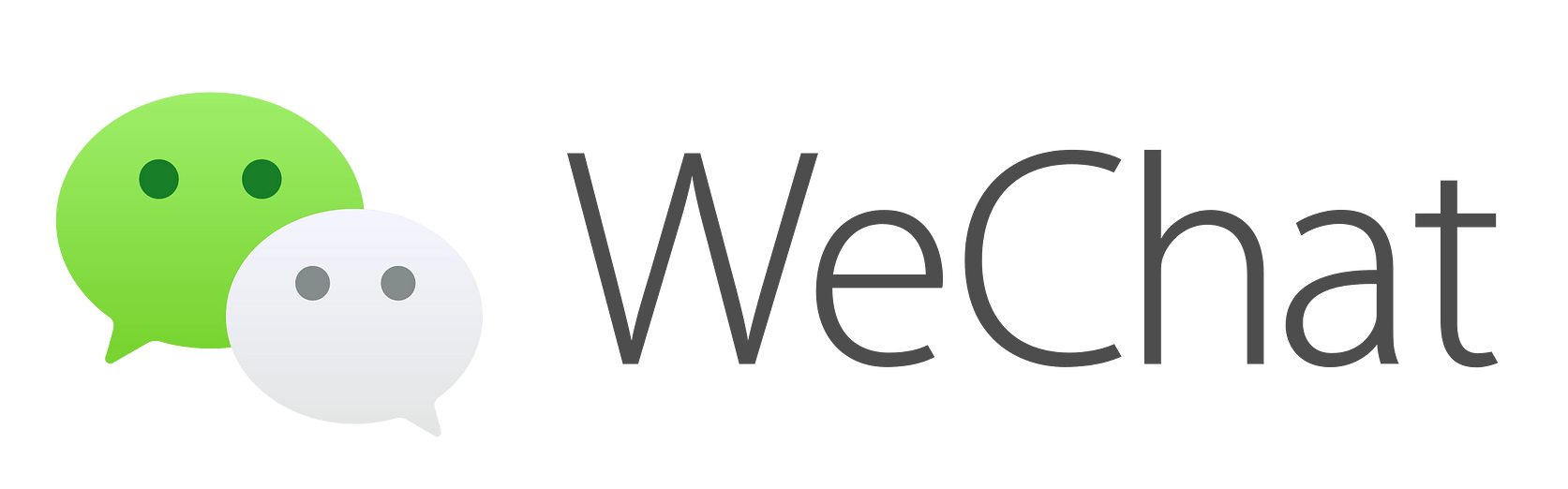}}}%
  \put(0,-60){\makebox[\paperwidth][l]{\hspace*{2.5cm}%
    \textcolor{wechatgreen}{\rule{\textwidth}{0.5pt}}}}%
}
\fi

\title{SkillHone: A Harness for Continual Agent Skill Evolution Through Persistent Decision History}

\ifemnlpsubmission
\author{Anonymous EMNLP submission}
\else
\author{Zhiwei Li \quad Yong Hu \\
  WeChat, Tencent Inc., China \\
  \texttt{zhiweili.jay@foxmail.com} \quad \texttt{rightyonghu@tencent.com} \\
  \textbf{Skills:} \url{https://github.com/Tencent/SkillHone} \\
  \textbf{Project Page:} \url{https://zwlijay.github.io/SkillHone-Project} \\
}
\fi

\begin{document}

\maketitle

\begin{abstract}
Agent skills extend language-model agents with task-specific procedures, scripts, and references, but the tasks and environments they target continually change.
Existing methods improve skills in bounded runs and retain only the final artifact, discarding the decision history that later agents need to interpret prior revisions, evaluations, and rejected alternatives.
We introduce SkillHone, a harness for continual agent skill evolution grounded in persistent decision history.
SkillHone pairs skill revisions with evaluation-side evidence that supplies practice feedback, recording structured histories of diagnoses, revisions, evidence, and outcomes.
Role-separated subagents run candidate skills on practice probes with redacted reporting and propose revisions informed by prior decisions, enabling cross-session refinement without rediscovering past rationale.
On deep-research benchmarks, SkillHone runs without a pre-integrated search stack and outperforms the commercially backed deep-research agent by 15.8 points on GAIA and 3.2 points on WebWalkerQA-EN, while also exceeding prior skill-evolution methods. We further deploy SkillHone on internal tool-mediated analysis scenarios, where it improves accuracy by an average of 18.8 points across seven settings.

\end{abstract}

\section{Introduction}

Skills have emerged as a central mechanism for extending LLM-based agents. A skill is a named, loadable bundle of task-specific procedures, scripts, references, and output conventions that an agent can invoke for a particular class of work~\citep{xu2026agent,ling2026agent}. In deployed agent systems such as Claude Code, Codex, and Hermes, skills serve as a primary unit of specialization. The development of a robust skill is never a one-shot event but an \textbf{ongoing process of creation and adaptation}. Even during its initial formation, a skill requires rigorous debugging and constant updates to broaden its scope or patch unhandled edge cases. As environments shift and APIs change, this creation-maintenance loop persists, making it essential that the rationale behind every revision remains preserved and maintainable throughout its existence. The central methodological question is therefore how later agents can continue improving a skill across many development sessions without losing the rationale behind earlier changes.

\begin{figure}[t]
\centering
\includegraphics[width=0.88\columnwidth]{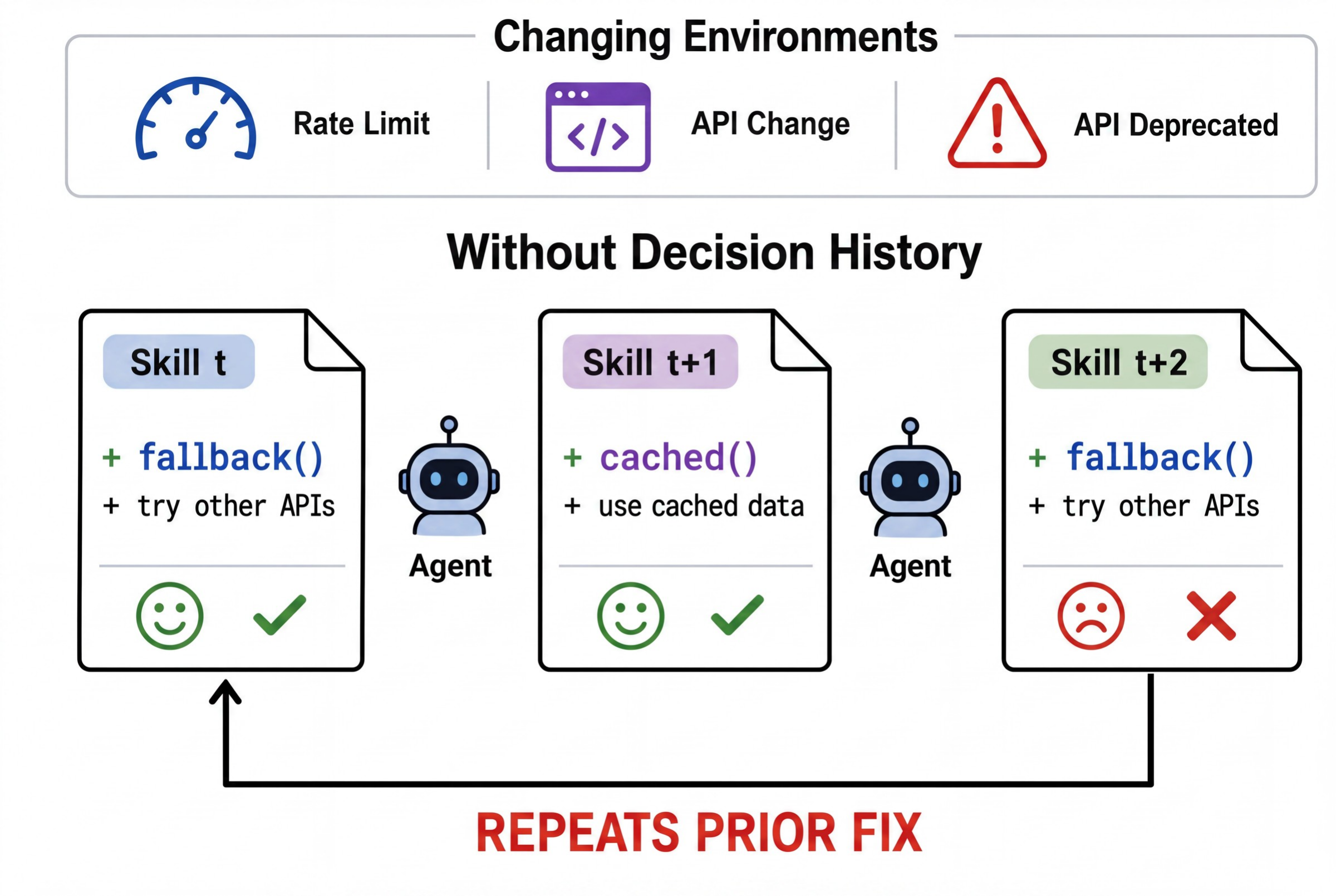}
\caption{
Context loss in artifact-only skill evolution: without decision history, a later agent may repeat a prior optimization and make future improvement harder.
}
\label{fig:teaser}
\end{figure}

Existing methods for improving agent skills can be viewed as two representative routes.
Synthesis-style construction methods such as Skill-Creator\footnote{\url{https://github.com/anthropics/skills/blob/main/skills/skill-creator}} turn a task description, examples, or supporting documents into a loadable skill and then hand the generated artifact to later sessions.
GEPA-style reflective optimization methods, including Hermes-Agent-Self-Evolution (Hermes-SE)\footnote{\url{https://github.com/NousResearch/hermes-agent-self-evolution}}, run iterative mutation, evaluation, and selection over skills, prompts, tool descriptions, and code using GEPA~\citep{agrawal2026gepa}.
These routes are useful for producing or improving a skill within a single optimization run, but they largely treat the output as a better artifact.

This artifact-centered view loses the decision context needed for continual maintenance.
A later agent may inherit the latest skill, but not the structured history explaining which diagnosis motivated a revision, which alternatives were rejected, what evaluation evidence was observed, and why an outcome was accepted.
The same problem applies to the evaluation evidence that supports skill improvement: when later agents rely on probes, validators, traces, or redacted reports, they also need to know which failure mode the evidence reflects and how it shaped prior decisions.
As illustrated in Figure~\ref{fig:teaser}, rate limits, API changes, and deprecations can arrive over time, and a later optimization step may repeat a previously successful repair that no longer matches the current failure mode.
Persistent decision history lets later optimization recover why each revision was made and avoid reintroducing obsolete fixes.
Without this context, a locally plausible repair can repeat old edits or make later optimization harder.

We introduce SkillHone, an agent-facing harness for continual skill evolution in skill-enabled agent frameworks.
SkillHone maintains the evolving skill together with evaluation assets that support its improvement, and treats skill revisions and evaluation evidence as decision-bearing.
For skill revisions, the history records the diagnosis, proposed revision, evaluation evidence, and outcome.
For evaluation evidence, it records which probes or reports were used, which failure mode they exposed, and how the evidence justified the outcome.
Thus, later agents inherit the skill, evidence, and rationale behind prior changes.

This history is accumulated from many local decisions made during diagnosis, revision, and evaluation.
SkillHone therefore uses role-bounded subagent dispatches, rather than a fixed multi-agent system, to generate these records as the loop runs: it only assumes an agent runtime that can launch scoped dispatches with separated permissions.
This makes the harness portable across agent runtimes such as Claude Code, Codex, and Hermes, while keeping optimization separated from unredacted evaluation assets.
Evaluation subagents can inspect probes and traces and run candidate skills, but return only redacted evidence to the optimization side.
Optimization subagents use this evidence and prior decision records to form diagnoses and draft skill revisions, without accessing hidden probe targets or validators.
Across sessions, the linked history of diagnoses, revisions, evidence, and outcomes allows later agents to continue skill evolution without re-deriving prior decisions.

We evaluate SkillHone on deep-research benchmarks in the raw open-web setting, where agents must organize retrieval without a pre-integrated search stack.
Against skill-evolution baselines and a deep-research agent with commercial retrieval, SkillHone obtains the strongest average results on GAIA and WebWalkerQA-EN under a fixed evaluation-time backbone; the resulting skill bundle also transfers to a different execution backbone (Claude Sonnet 4.6) without additional optimization, indicating that the gain reflects the skill procedure rather than fitting to one model.
We also deploy SkillHone on recurring internal tool-mediated analysis scenarios, where it improves the underlying tool-use skills.

\noindent\textbf{Contributions.}
\begin{enumerate}[leftmargin=1.4em,itemsep=2pt,topsep=2pt]
\item We identify a gap between artifact-centered skill improvement and continual skill maintenance.
  Long-lived skills require persistent decision history, rather than a single optimized artifact, so later agents can understand skill revisions together with the evaluation evidence that supported them.

\item We introduce SkillHone, a general agent-facing harness for continual agent skill evolution.
  It records diagnoses, revisions, evidence, and outcomes under separated roles.
  
\item We show that SkillHone improves skill development across public benchmarks and deployments.
  
\end{enumerate}

\section{Method}
\label{sec:method}

\begin{figure*}[t]
  \centering
  \includegraphics[width=0.98\textwidth]{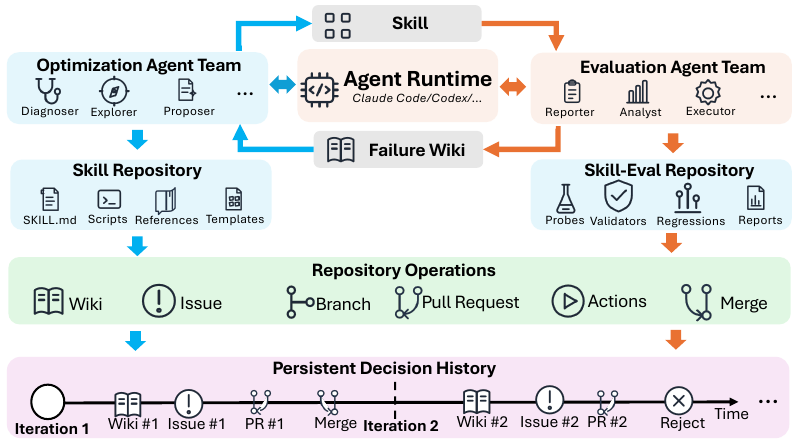}
  \caption{
  SkillHone architecture. At each development step, an agent runtime dynamically dispatches fresh, role-bounded optimization and evaluation subagents on demand. Optimization dispatches maintain the skill repository and propose diagnosis-guided revisions. Evaluation dispatches run candidate skills against the skill-evaluation repository and export redacted problem reports, without exposing probe targets, validators, or traces. Routed evidence, repository operations, and recorded outcomes form a persistent decision history that later agents reuse for continual skill evolution.
  }
  \label{fig:framework}
\end{figure*}

\subsection{Problem Setup}
\label{sec:problem-setup}

In this paper, a skill is a loadable artifact that packages the instructions, procedures, references, and output conventions an agent uses for a class of tasks. SkillHone keeps the model backend and external tools fixed; its goal is to help agents maintain and improve this skill artifact across repeated development sessions.

SkillHone maintains two linked repositories. The skill repository stores the skill bundle being revised. The skill-evaluation repository stores practice probes, oracle targets, validators, traces, redacted reports, and other evaluation assets that provide feedback for skill improvement. The optimization side observes only redacted reports, not unredacted targets, validators, or traces. Accepted revisions produce a sequence of skills $S_0,\ldots,S_T$, guided by practice feedback and the persistent decision history that links diagnoses, candidate revisions, evaluations, and outcomes.

\subsection{SkillHone Harness}
\label{sec:harness}

As shown in Figure~\ref{fig:framework}, SkillHone separates each development step into optimization and evaluation dispatches. These dispatches are role-bounded so that diagnoses, revisions, redacted evidence, and outcomes can be recorded as auditable decision history. Appendix~\ref{app:subagent-roles} (Table~\ref{tab:roles}) lists the recurring dispatch patterns and their permission boundaries.

We formalize the harness as
\[
\mathcal{M}=(\mathcal{T}_{\mathrm{opt}},\mathcal{T}_{\mathrm{eval}},\mathcal{D}),
\]
where $\mathcal{T}_{\mathrm{opt}}$ is the Optimization Agent Team, $\mathcal{T}_{\mathrm{eval}}$ is the Evaluation Agent Team, and $\mathcal{D}$ is the runtime dispatcher.

The optimization side maintains the skill repository, which contains the current \texttt{SKILL.md}, scripts, references, templates, and other skill assets. Optimization dispatches can inspect the current skill, read prior decision records and failure summaries, form diagnoses, propose typed revisions, review candidate changes, and record accepted outcomes. Typical dispatches include diagnosing failures from redacted reports, integrating prior resources, proposing scoped revisions, and drafting patches to $S_t$.

The evaluation side operates on the skill-evaluation repository, which contains probes, validation suites, regression sets, traces, and evaluation reports. Evaluation dispatches can inspect oracle targets and traces, run the current skill on probe items, analyze failures, and produce redacted reports. They cannot write to the skill repository. Conversely, the optimization side can change the skill but cannot access unredacted probe targets, validators, or execution traces. This separation prevents practice feedback from becoming a direct memorization target.

The dispatcher is a message router rather than a decision authority. At each step, it routes redacted evidence and review requests between the two sides and records accepted outcomes. Acceptance decisions are made using the evidence available to the appropriate side, keeping the decision history tied to the same permission boundary that produced it.

\subsection{Persistent Decision History}
\label{sec:decision-history}

A central design goal of SkillHone is to make skill evolution persistent across sessions. A later agent should not receive only the latest skill snapshot. It should also inherit the structured record of why the skill changed, which alternatives were considered, how candidates were evaluated, and which outcomes were accepted or rejected.

SkillHone represents each development step as a decision record
\[
h_t=(q_t,r_t,e_t,o_t),
\]
where $q_t$ is the diagnosis, $r_t$ is the candidate revision, $e_t$ is the redacted evaluation evidence, and $o_t$ is the outcome. The outcome may accept the revision, reject it, request further changes, or defer the diagnosis. The decision history up to iteration $t$ is $\mathcal{H}_t=\{h_1,\ldots,h_t\}$.

The decision record is more than a version diff. A diff shows how files changed; a decision record links that change to the problem it targeted, the evidence used to evaluate it, and the final decision. This distinction is important for continual evolution. When the environment changes, later subagents can inspect $\mathcal{H}_t$ to determine whether a failure is new, whether a similar fix was already attempted, and why a previous alternative was rejected.

Given a new redacted report $\tilde{E}_t$, optimization subagents retrieve relevant prior records from $\mathcal{H}_{<t}$. They may revisit an old diagnosis, revise a rejected patch under new evidence, or avoid a redundant edit that has already failed. Thus the decision history acts as a persistent memory for the evolution process itself. It enables agents to audit prior changes and continue skill improvement without re-deriving the same diagnoses.

\begin{algorithm}[t]
  \small
  \caption{SkillHone}
  \label{alg:skillhone}
  \begin{algorithmic}[1]
  \Require Initial skill $S_0$, skill repo $\mathcal{R}_s$, eval repo $\mathcal{R}_e$
  \State $\mathcal{H}_0 \leftarrow \emptyset$
  \For{$t=0,\ldots,T-1$}
      \State Eval dispatches run $A(\cdot;S_t)$ on $\mathcal{R}_e$ and export $\tilde{E}_t$
      \State Opt dispatches retrieve $\mathcal{H}_t$, then propose $(q_t,r_t)$
      \State Eval dispatches return redacted evidence $e_t$ for $r_t$
      \State Decide $o_t \in \{\textsc{accept},\textsc{revise},\textsc{reject},\textsc{defer}\}$
      \State $S_{t+1}\leftarrow \textsc{Apply}(S_t,r_t)$ if accepted, else $S_t$
      \State $\mathcal{H}_{t+1}\leftarrow \mathcal{H}_t\cup\{(q_t,r_t,e_t,o_t)\}$
  \EndFor
  \State \Return $S_T,\mathcal{H}_T$
  \end{algorithmic}
  \end{algorithm}

  \section{Experiments}
  \label{sec:experiments}

\begin{table*}[!t]
  \centering
  \small
  \setlength{\tabcolsep}{3.5pt}
  \begin{tabular}{ll|cccc|cccc}
    \toprule
                  & & \multicolumn{4}{c|}{\textbf{GAIA}} & \multicolumn{4}{c}{\textbf{WebWalkerQA-EN}} \\
    \cmidrule(lr){3-6} \cmidrule(lr){7-10}
    \textbf{Setting} & \textbf{System} & L1 & L2 & L3 & Avg. & Easy & Med. & Hard & Avg. \\
    \midrule
    Curated search & deep-research agent & 61.9 & 47.0 & 26.3 & 48.8 & \textbf{58.5} & 62.3 & 67.1 & 63.2 \\
    \midrule
    \multirow{4}{*}{Raw open-web} & Existing-Skills & 64.3 & 33.3 & 21.1 & 41.7 & 51.2 & 48.5 & 52.6 & 50.2 \\
    & Skill-Creator & 64.3 & 37.9 & 21.1 & 44.1 & 36.6 & 36.9 & 40.8 & 38.1 \\
    & Hermes-SE & 73.8 & 40.9 & \textbf{31.6} & 50.4 & 53.7 & 51.5 & 55.3 & 53.0 \\
    & SkillHone (Ours) & \textbf{76.2} & \textbf{66.7} & \textbf{31.6} & \textbf{64.6} & 53.7 & \textbf{69.2} & \textbf{68.4} & \textbf{66.4} \\
    \bottomrule
    \end{tabular}
    \caption{
      Main results on GAIA and WebWalkerQA-EN.
      Bold marks the best score in each column.
    }
  \label{tab:main-results}
  \end{table*}

  We evaluate SkillHone on public deep-research benchmarks against existing skill construction and optimization methods under a fixed execution-time agent backbone, examine cross-backbone transfer of the resulting skill bundle, and report a deployment study on recurring internal tool-mediated analysis scenarios.

\subsection{Experimental Setup}

We instantiate this comparison in deep research because information seeking is one of the most broadly used agent workloads and naturally benefits from reusable skills: tasks require agents to combine search, navigation, extraction, verification, and recovery across changing web sources, so external resources are rate-limited, fail intermittently, and change format, requiring agents to diagnose failures, revise skills, and preserve the evaluation evidence behind those revisions.

We use two open-domain benchmarks: the text-only validation subset of GAIA~\citep{mialon2024gaia} and the English split of WebWalkerQA~\citep{wu2025webwalker}.
GAIA emphasizes multi-step information seeking and factual grounding across heterogeneous sources, while WebWalkerQA-EN focuses on web navigation and information extraction.

We compare two evaluation settings:
\begin{itemize}[leftmargin=1.4em,itemsep=2pt,topsep=2pt]
    \item \textbf{Curated search}: a deep-research agent uses commercial retrieval services and does not maintain its own search stack.
    \item \textbf{Raw open-web}: agents receive no pre-integrated search tools. They operate through a general agent runtime, public web access, and portable skill bundles that organize search, parsing, extraction, and recovery procedures. Because these skills rely on public pages and free or community search interfaces, they must handle rate limits, unavailable endpoints, and changing page formats without a managed retrieval service.
\end{itemize}

We separate the controller that optimizes skills from the backbone that executes and evaluates them.
Claude Opus 4.6\footnote{\url{https://www.anthropic.com/news/claude-opus-4-6}} serves as the development-time controller that reads feedback, forms diagnoses, and revises skill bundles.
Qwen3.6-35B-A3B~\citep{qwen36_35b_a3b} serves as the execution and evaluation backbone for running skills during development and for producing the final benchmark scores.

To drive skill evolution, we sample 20 items from the publicly released WebShaper dataset~\citep{tao2025webshaper}, which is disjoint from GAIA and WebWalkerQA-EN, and adapt them to our agent execution environment.
These items serve as the shared development data for all skill-evolution baselines, with each method using them according to its own optimization procedure.

\subsection{Baselines}

The raw open-web comparison starts from a shared existing-skill pool.
We construct this pool from search-related community skills selected by an agent from ClawHub and SkillHub; Appendix~\ref{app:existing-skills-list} lists the selected skills.
The pool is provided to all comparable raw open-web systems for a fair comparison; the systems differ in whether and how they adapt these starting skills.

The raw open-web comparison contains one direct-use setting and three skill-development systems:
\begin{itemize}
    \item \textbf{Existing-Skills}: directly loads the skills as a portable skill bundle, without further optimization.
    \item \textbf{Skill-Creator}: applies built-in iterative skill optimization.
    \item \textbf{Hermes-SE}: applies reflective skill optimization.
    \item \textbf{SkillHone}: runs the full harness with role-separated subagents, redacted evaluation feedback, and persistent decision history.
\end{itemize}

All raw open-web systems produce the same portable skill-bundle interface: a \texttt{SKILL.md} entry file with optional scripts, references, templates, and supporting resources.
They differ only in how the bundle is developed.

  \subsection{Main Results}

  Table~\ref{tab:main-results} reports the final benchmark results.
  SkillHone achieves the best average score on both benchmarks, reaching $64.6\%$ on GAIA and $66.4\%$ on WebWalkerQA-EN, $15.8$/$3.2$ points above the deep-research agent despite not using pre-integrated search tools at evaluation time.
  The gains concentrate on harder splits that require deciding how to search, extracting evidence, verifying findings, and recovering from failed attempts.

  Within the raw open-web setting, SkillHone also outperforms the shared-pool skill-development baselines: $+20.5$/$+28.3$ points over Skill-Creator and $+14.2$/$+13.4$ over Hermes-SE on GAIA/WebWalkerQA-EN.
  The gap to Existing-Skills shows that exposing the agent to relevant skills is not sufficient: SkillHone turns them into a task procedure for when to search broadly, follow a source, verify an answer, and recover from a failed path, so the main gains come from organization rather than components alone.

\begin{figure}[t]
\centering
\includegraphics[width=\columnwidth]{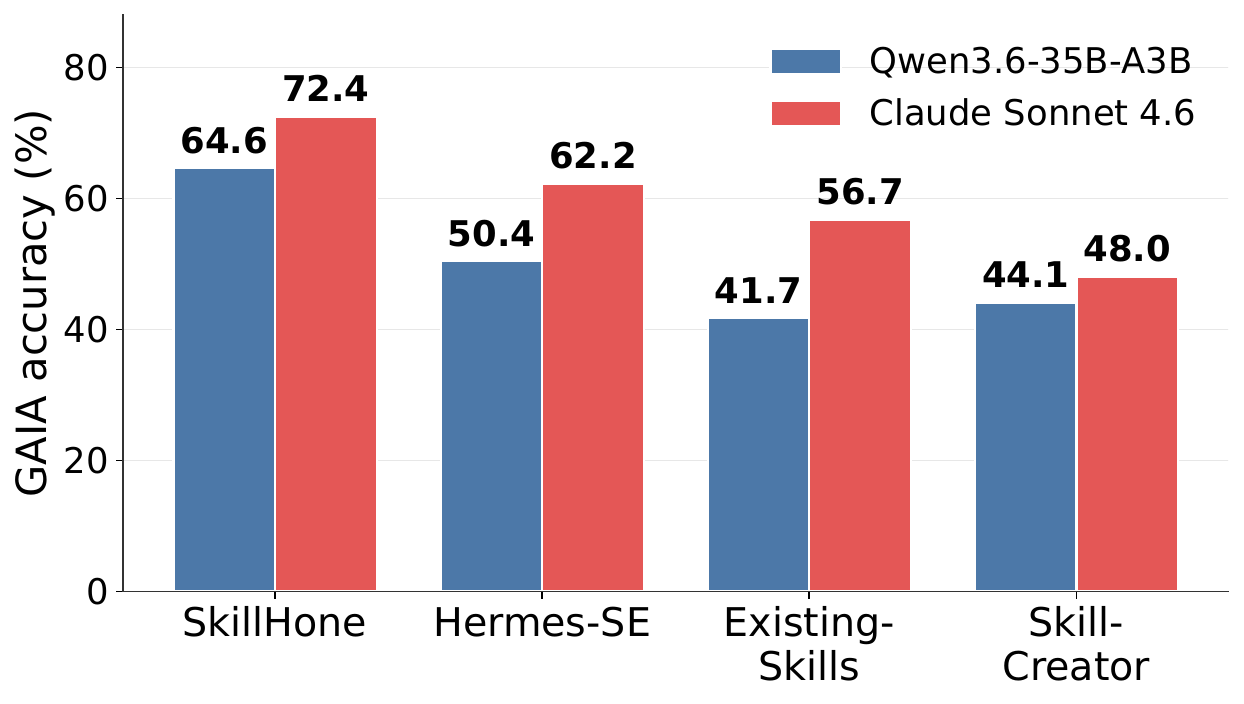}
\caption{
GAIA accuracy under the development backbone (Qwen3.6-35B-A3B) and the transfer backbone (Claude Sonnet 4.6). Each skill bundle is run directly on Claude Sonnet 4.6 without additional optimization.
}
\label{fig:gaia-transfer-sonnet}
\end{figure}

  Figure~\ref{fig:gaia-transfer-sonnet} further evaluates transfer to Claude Sonnet 4.6\footnote{\url{https://www.anthropic.com/news/claude-sonnet-4-6}}: without additional optimization on this backbone, SkillHone reaches $72.4\%$ on GAIA, $10.2$/$15.7$/$24.4$ points above Hermes-SE/Existing-Skills/Skill-Creator.
Appendix~\ref{app:trajectory} (Figure~\ref{fig:trajectory}) compares the SkillHone and Hermes-SE optimization trajectories on the same practice probes, illustrating how scoped reverts under persistent decision history differ from whole-candidate acceptance under a scalar signal.
Appendix~\ref{app:optimization-cost} reports development cost and cost-efficiency metrics.

\subsection{Ablation}
\label{subsec:ablation}

To isolate the two core mechanisms, we ablate each under the raw open-web setting and the same Qwen3.6-35B-A3B execution backbone used above.
\textbf{No decision history} keeps role-separated subagents but starts each step from the latest skill artifact alone, optimizing on a scalar signal as in reflective baselines.
\textbf{No role separation} keeps decision history but lets a single subagent access the skill repository and unredacted evaluation assets jointly.

Table~\ref{tab:ablation} shows that decision history accounts for the larger share: removing it drops GAIA/WebWalkerQA-EN by $13.4$/$10.9$ points, versus $6.4$/$5.3$ points for role separation.

\begin{table}[t]
  \centering
  \small
  \setlength{\tabcolsep}{6pt}
  \begin{tabular}{lcc}
    \toprule
     & \textbf{GAIA} & \textbf{WebWalkerQA-EN} \\
    \midrule
    SkillHone (full)            & $64.6$ & $66.4$ \\
    \quad w/o decision history  & $51.2$ & $55.5$ \\
    \quad w/o role separation   & $58.2$ & $61.1$ \\
    \bottomrule
  \end{tabular}
  \caption{
    Ablation of SkillHone's two core mechanisms.
  }
  \label{tab:ablation}
\end{table}

\subsection{Deployment Study}
\label{subsec:deployment}

We further evaluate SkillHone on internal tool-mediated analysis scenarios to test whether it improves skills used for recurring analytical requests beyond public web benchmarks.
We select several recurring scenarios and their corresponding seed skills from an internal workflow, then apply SkillHone to improve each skill on a fixed exact-match evaluation set per scenario (Appendix~\ref{app:deployment-protocol}).

Table~\ref{tab:deployment-results} summarizes the selected scenarios.
SkillHone improves six of the seven selected skills over their seeded versions, with the largest gains on counting, aggregation, structure parsing, and span retrieval.
The remaining scenario, list filtering, is stable at the aggregate level.
Across all listed scenarios, SkillHone improves accuracy by an average of $18.8$ points, with the largest gains where the initial procedure leaves reusable analysis steps underspecified.

\begin{table}[t]
  \centering
  \small
  \setlength{\tabcolsep}{6pt}
  \begin{tabular}{lr}
    \toprule
    \textbf{Scenario} & $\Delta$ \\
    \midrule
    Counting           & $+30.0$ \\
    Aggregation        & $+26.3$ \\
    Structure parsing  & $+25.0$ \\
    Density estimation & $+23.1$ \\
    Span retrieval     & $+21.5$ \\
    Filtered ranking   & $+5.9$  \\
    List filtering     & $+0.0$ \\
    \midrule
    \textbf{Avg.}                & $+18.8$ \\
    \bottomrule
  \end{tabular}
  \caption{
    Deployment study on recurring internal tool-mediated analysis scenarios. Optimized minus seeded accuracy (pp); the final row averages the listed scenarios.
  }
  \label{tab:deployment-results}
\end{table}

\section{Related Work}
\label{sec:related}

\paragraph{Agent skills and skill acquisition.}
The recent abstraction of agent skills bundles task-specific procedures, scripts, references, and conventions into reusable artifacts that LLM agents can load on demand~\citep{xu2026agent,ling2026agent}.
Benchmarks such as~\citet{li2026skillsbench} and \citet{han2026swe} measure when skills help across general tasks and software-engineering settings, while skill-acquisition methods study automated discovery from exploration and feedback~\citep{yang2025automated}, reinforcement-learning-based evolution~\citep{vishe2026skill,xia2026skillrl}, and domain-specific construction~\citep{liu2026skillforge}.
These works focus on producing or evaluating skills as outputs; SkillHone instead preserves the optimization context---failure modes, candidate revisions, and outcomes---so the next agent inherits the decision history that makes future improvement possible.

\paragraph{Optimization of prompts, systems, and skills.}
Within language-system optimization, prompt-only methods use LLMs to propose and select improved instructions~\citep{zhou2022large,yang2024large}; Reflexion-style agents use verbal feedback to improve behavior across attempts~\citep{shinn2023reflexion}; DSPy frames LM pipelines as declarative programs that can be compiled and optimized end-to-end~\citep{khattab2023dspy}.
More closely related, GEPA reflectively evolves components of a language system~\citep{agrawal2026gepa}, and Hermes-SE applies this style across skills, prompts, tool descriptions, and code.
These methods optimize within a single run over a scalar signal; SkillHone lets later agents continue from where prior runs left off, audit what has been tried, and avoid redundant edits.

\paragraph{Multi-agent collaboration and agent development workflows.}
Multi-agent LLM systems study how role specialization can support collaborative problem solving~\citep{talebirad2023multi,guo2024large,wang2024survey,tran2025multi}.
MetaGPT organizes agents into specialized roles for producing software-like artifacts~\citep{hong2024metagpt}, and SWE-style benchmarks~\citep{jimenez2024swe,yang2024swe,xia2025live} study LLM agents in environments where tasks are naturally expressed as diagnosis, code modification, and testing.
SkillHone reuses role separation for a different purpose: optimization subagents propose skill revisions while evaluation subagents test candidates and return redacted evidence under a permission-bounded dispatcher, accumulating agent-readable decision history while reducing leakage from evaluation assets to the optimization side.

\section{Conclusion}

We introduced SkillHone, a harness that evolves skills via role-separated subagents, redacted practice feedback, and persistent decision histories linking problems, revisions, evidence, and outcomes.
On GAIA and WebWalkerQA-EN, SkillHone exceeds iterative and reflective skill-development baselines and the deep-research agent, and improves seeded skills by $18.8$ points on average across seven recurring internal scenarios.
The skill bundle transfers to a different execution backbone without further optimization, suggesting the gain comes from the skill procedure rather than from fitting one model.
Decision history lets later agents improve a skill without re-deriving prior decisions.

\section*{Limitations}

SkillHone currently evolves a single skill in isolation; jointly evolving multiple interdependent skills, including coordination across shared resources and overlapping failure modes, is not addressed in this work.

\section*{Acknowledgments}

This work was conducted while Zhiwei Li was interning at WeChat AI, Tencent Inc., China.

\bibliography{custom}

\appendix

\section{Existing Community Skills}
\label{app:existing-skills-list}

The shared existing-skill pool is selected by an agent from search-related community skills on ClawHub and SkillHub:
\begin{itemize}[leftmargin=1.4em,itemsep=1pt,topsep=2pt]
    \item \texttt{web-pilot}
    \item \texttt{web-content-fetcher}
    \item \texttt{scholar-search}
    \item \texttt{query-dbpedia}
    \item \texttt{multi-search-engine}
    \item \texttt{literature-review}
    \item \texttt{geepers-data}
    \item \texttt{duckduckgo-websearch}
    \item \texttt{deep-research-pro}
    \item \texttt{ddgs-search}
\end{itemize}

\section{Repository Workflow Example}
\label{app:issue-pr-evolution}

In our implementation, SkillHone realizes persistent decision history through a GitHub-style repository workflow.
Issues record diagnosed failure modes, pull requests contain proposed skill revisions, and merge or rejection decisions record the accepted outcome.
This is one concrete interface for making the evolution process auditable, rather than a required interface for SkillHone.

Table~\ref{tab:issue-pr-evolution} gives the repository records from one deep-research skill run.
The five rows correspond to the five non-seed SkillHone iterations shown in Figure~\ref{fig:trajectory}.
The table highlights that SkillHone optimizes the whole skill bundle, including instructions, scripts, and references, rather than only rewriting a single \texttt{SKILL.md} body as in the Hermes-SE comparison.

\begin{table*}[t]
\centering
\small
\setlength{\tabcolsep}{4pt}
\begin{tabular}{>{\raggedright\arraybackslash}p{0.04\textwidth}
                >{\raggedright\arraybackslash}p{0.29\textwidth}
                >{\raggedright\arraybackslash}p{0.19\textwidth}
                >{\raggedright\arraybackslash}p{0.38\textwidth}}
\toprule
\textbf{Iter.} & \textbf{Issues} & \textbf{PRs} & \textbf{Files changed} \\
\midrule
1 &
\#1 DuckDuckGo search, error handling, and time management &
\#2 DuckDuckGo search, error handling, and time management &
\texttt{SKILL.md} (+29/-13), \texttt{scripts/ddg\_search.sh} (+25/-0), \texttt{scripts/web\_search.sh} (+25/-7), \texttt{scripts/wiki\_page.sh} (+15/-3), \texttt{scripts/wiki\_search.sh} (+15/-3), \texttt{scripts/wiki\_section.sh} (+39/-9), \texttt{scripts/wikidata\_lookup.sh} (+52/-0) \\
\midrule
2 &
\#3 real web search, script fixes, write-early strategy &
\#4 real web search, script fixes, write-early strategy &
\texttt{SKILL.md} (+33/-48), \texttt{scripts/ddg\_search.py} (+53/-0), \texttt{scripts/fetch\_url\_jina.sh} (+25/-0), \texttt{scripts/web\_search.sh} (+6/-6), \texttt{scripts/wiki\_page.sh} (+3/-3), \texttt{scripts/wiki\_search.sh} (+3/-3), \texttt{scripts/wiki\_section.sh} (+7/-7) \\
\midrule
3 &
\#11 merge best of iter1 and iter2 with lighter verification and stricter budget &
\#12 merge best of iter1 and iter2 &
\texttt{SKILL.md} (+33/-40) \\
\midrule
4 &
\#15 name-format issue and tighter budget &
\#16 name-format rules and Wikidata SPARQL &
\texttt{SKILL.md} (+13/-6), \texttt{scripts/wikidata\_sparql.py} (+59/-0) \\
\midrule
5 &
\#19 revert budget to 20 while keeping name rules and SPARQL &
\#21 revert budget to 20, keep name rules and SPARQL &
\texttt{SKILL.md} (+45/-46) \\
\bottomrule
\end{tabular}
\caption{
Repository workflow records for one deep-research skill run.
Line counts aggregate the changed files in the PRs listed for each iteration.
}
\label{tab:issue-pr-evolution}
\end{table*}

\section{Optimization Trajectory}
\label{app:trajectory}

\begin{figure*}[t]
\centering
\includegraphics[width=0.96\textwidth]{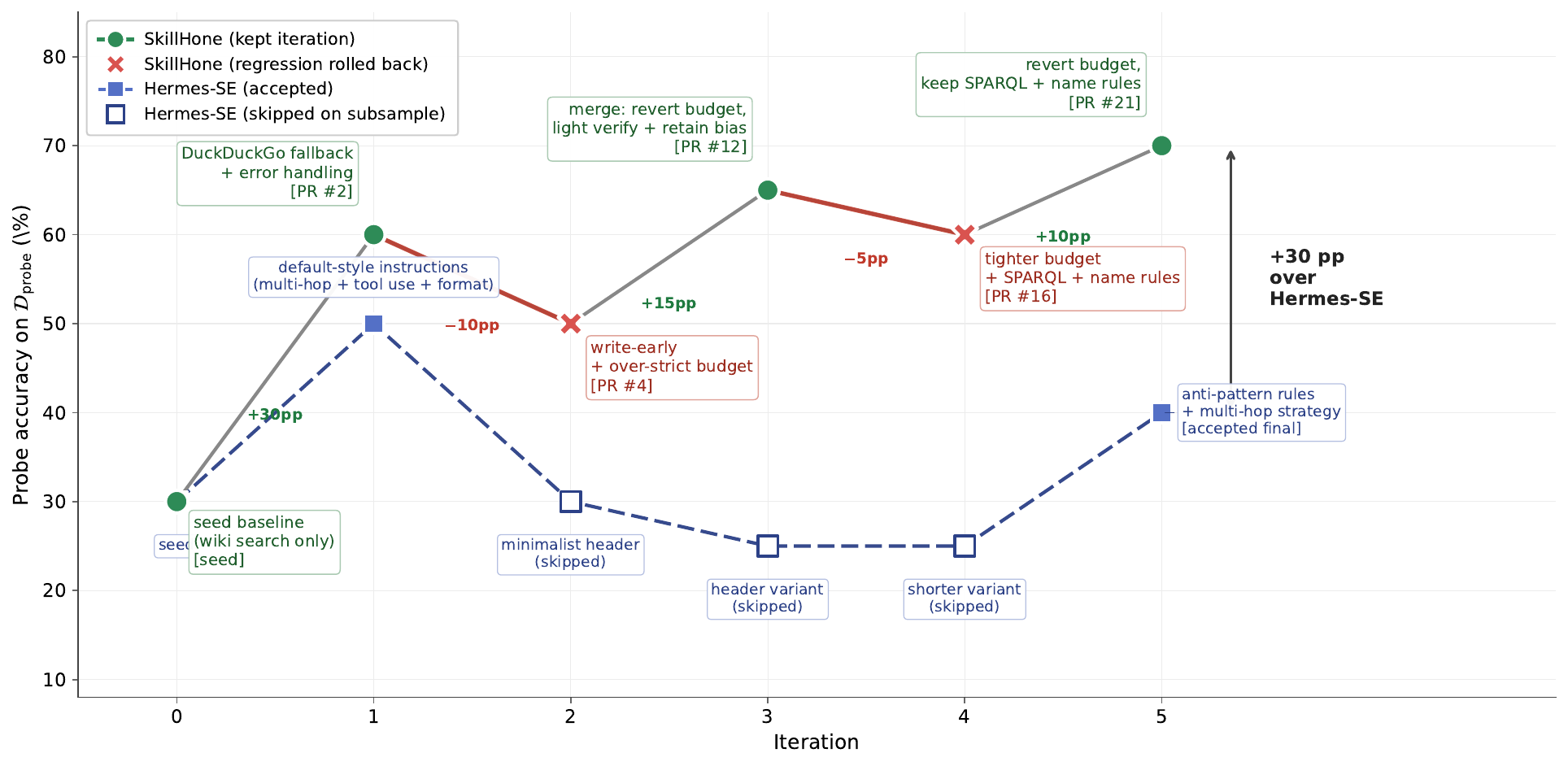}
\caption{
Probe accuracy trajectories for SkillHone and Hermes-SE across five optimization iterations starting from a shared seed skill.
SkillHone points are labeled with the accepted revision that introduced each change.
When a change regresses performance, the following iteration can target the offending part while preserving useful edits.
Hermes-SE optimizes a single \texttt{SKILL.md} body and accepts or rejects whole candidates under a scalar validation signal.
The final benchmarks are not available to either system during optimization.
}
\label{fig:trajectory}
\end{figure*}

Figure~\ref{fig:trajectory} compares one SkillHone run with one Hermes-SE run on the same practice probes.
Both runs start from the same seed skill.
SkillHone improves from $30\%$ to $70\%$ while recovering from two regressed revisions through targeted follow-up edits.
Hermes-SE instead accepts or skips whole prompt candidates under a scalar validation signal.
This illustrates the operational value of persistent decision history: later revisions can target the offending part of a change while retaining useful edits.

\section{Optimization Cost and Efficiency}
\label{app:optimization-cost}

Table~\ref{tab:optimization-cost} compares the development cost and cost efficiency of Hermes-SE and SkillHone.
Gain over Skill-Creator is the mean of the GAIA and WebWalkerQA-EN benchmark-average improvements; gain per dollar divides this mean by total cost, and cost per gain point is its reciprocal.
SkillHone obtains a larger average gain with fewer model calls and fewer output tokens, while using substantially more input tokens.

\begin{table}[t]
  \centering
  \small
  \setlength{\tabcolsep}{4pt}
  \begin{tabular}{lrr}
    \toprule
    \textbf{Metric} & \textbf{Hermes-SE} & \textbf{SkillHone} \\
    \midrule
    \multicolumn{3}{l}{\textit{Gain and cost efficiency}} \\
    Gain over Skill-Creator & $+10.6$ & $\mathbf{+24.4}$ \\
    Gain per dollar & $0.420$ & $\mathbf{0.845}$ \\
    Cost per gain point & $2.38$ & $\mathbf{1.18}$ \\
    \midrule
    \multicolumn{3}{l}{\textit{Development resources}} \\
    Development cost & $25.22$ & $28.89$ \\
    Model calls & $690$ & $627$ \\
    Input tokens & $1.48$M & $39.9$M \\
    Output tokens & $0.78$M & $0.21$M \\
    \bottomrule
  \end{tabular}
  \caption{Development cost and efficiency of Hermes-SE and SkillHone. Gain is measured in percentage points (pp), gain per dollar in pp/USD, cost per gain point in USD/pp, and development cost in USD. Bold marks the better gain or cost-efficiency value; raw resource usage is not ranked.}
  \label{tab:optimization-cost}
\end{table}

\section{Evaluation Repository and Redaction}
\label{app:evaluation-repository}

SkillHone uses a skill-evaluation repository for practice feedback during skill development.
The repository stores probe task instances, oracle targets, validators, provenance metadata, and execution traces.
Probe items may come from existing sources such as human-written cases, historical failures, or environment-derived checks.
It only assumes that evaluation subagents can run the current skill against available probe items and export redacted reports to the optimization side.

For each probe item, we denote the task instance by $x_k$, the oracle target by $y_k$, the validator by $\phi_k$, and the provenance metadata by $\pi_k$.
Running the current skill produces
\[
m_k=\phi_k(A(x_k;S_t), y_k),
\]
where $m_k$ is the validation outcome.
The optimization side receives only a redacted problem report
\[
\tilde{E}_t=\mathrm{Redact}(\{x_k,y_k,\phi_k,\pi_k,m_k\}_{k=1}^{K}),
\]
which summarizes failure modes, aggregate outcomes, and diagnostic hypotheses without exposing unredacted targets, validators, or traces.
Final benchmark evaluation remains outside the optimization loop.

\section{Subagent Dispatch Patterns and Permissions}
\label{app:subagent-roles}

The roles in Table~\ref{tab:roles} are not fixed identities that SkillHone preconfigures in advance.
They are recurring dispatch patterns observed when the runtime dispatcher runs the development loop.
Each subagent is created on demand, assigned to a permission-bounded team, and granted only the actions needed for that dispatch.

The team boundary is the structural mechanism.
Optimization subagents can write to the skill repository but cannot access unredacted probe targets, validators, or traces.
Evaluation subagents can inspect probes and traces but cannot write to the skill repository.
The runtime dispatcher creates subagents and routes artifacts, but it does not directly edit either repository.

\begin{table*}[!t]
\centering
\small
\setlength{\tabcolsep}{5pt}
\renewcommand{\arraystretch}{1.2}
\begin{tabular}{llp{8.4cm}}
\toprule
\textbf{Team} & \textbf{Dispatch pattern} & \textbf{Permission boundary} \\
\midrule
Optimization
& Proposer
& Reads redacted reports $\tilde{E}_{<t}$ and decision history $\mathcal{H}_{<t}$; writes diagnoses; no access to probe targets or validators. \\

Optimization
& Explorer
& Searches external resources for reusable patterns when a revision needs them; no access to probe targets or validators. \\

Optimization
& Developer
& Edits $S_t$ and proposes typed revisions; no access to unredacted evaluation assets. \\

Optimization
& Reviewer
& Reviews pending skill changes using redacted evidence and prior decision records; no access to probe targets or validators. \\

Optimization
& Decider
& Accepts or rejects candidate revisions according to recorded evidence; no access to unredacted evaluation assets. \\

\midrule
Evaluation
& Executor
& Runs $A(\cdot;S_t)$ on probe items; may inspect oracle targets, validators, and traces; no skill-repository write access. \\

Evaluation
& Diagnoser
& Analyzes outcomes and traces; updates evaluation diagnostics; no skill-repository write access. \\

Evaluation
& Reporter
& Produces redacted problem reports $\tilde{E}_t$ for the optimization side; no revision-decision authority. \\

Evaluation
& Auditor
& Checks probe metadata, traces, and redacted reports for evaluation-side consistency; no skill-repository write access. \\

\midrule
Runtime
& Dispatcher
& Creates subagents and routes artifacts between repositories; no direct skill edits and no access to unredacted probe targets. \\
\bottomrule
\end{tabular}
\caption{
Recurring subagent dispatch patterns in SkillHone.
The role column describes the function performed by a dispatched subagent, while the team column defines its permission boundary.
This separation makes the optimizer and evaluator split structural rather than prompt-imposed.
}
\label{tab:roles}
\end{table*}

\section{Deployment Evaluation Protocol}
\label{app:deployment-protocol}

Each scenario in Table~\ref{tab:deployment-results} is a recurring tool-mediated analysis task.
The per-scenario evaluation set consists of representative de-identified requests, with reference answers produced by LLM annotation and verified by human annotators; disagreements are resolved on review.
Accuracy is exact-match against the verified reference.
The seeded and SkillHone-optimized skills run on the same fixed evaluation set under identical runtime configuration, so the reported $\Delta$ reflects skill-level changes rather than evaluation-set drift.

\end{document}